\documentclass{article}
\usepackage{spconf,amsmath,graphicx}

\usepackage{makecell}
\usepackage[table]{xcolor}
\usepackage{multirow}
\usepackage{amsfonts}

\def\eg{\emph{e.g}.} 
\def\ie{\emph{i.e}.} 
 
\def\etc{\emph{etc}.} 
 
\def\etal{\emph{et al}.}
\def\ourmodel{GLCNet}

\newcommand{\figref}[1]{Fig.~\ref{#1}}
\newcommand{\tabref}[1]{Tab.~\ref{#1}}

\definecolor{mygray}{gray}{.92}

\makeatletter
\newcommand{\thickhline}{%
    \noalign {\ifnum 0=`}\fi \hrule height 1pt
    \futurelet \reserved@a \@xhline
}

\title{MovieNet-PS: A Large-Scale Person Search Dataset in the Wild}
\name{Jie Qin$^1$, Peng Zheng$^1$, Yichao Yan$^{2*}$, Rong Quan$^{1*}$\thanks{* Corresponding authors: Yichao Yan and Rong Quan.}, Xiaogang Cheng$^3$, Bingbing Ni$^2$}
\address{$^1$Nanjing University of Aeronautics and Astronautics\\
$^2$Shanghai Jiao Tong University~~
$^3$Nanjing University of Posts and Telecommunications}

\begin{document}
%
\maketitle
\begin{abstract}
Person search (PS) aims to jointly localize and identify a query person from natural, uncropped images. Existing works unintentionally adopt pedestrians (with similar poses and unchanging clothing) as the query and restrict the application scenarios in surveillance. This is due to that most PS datasets are collected from surveillance cameras with a limited diversity of views, scenes, appearances, \etc{} In this paper, we study a more general and realistic task in the wild, where we aim to search target persons with a much higher degree of diversity. To this end, we introduce a new PS dataset, namely MovieNet-PS, based on an existing large-scale movie dataset. MovieNet-PS is currently the largest and most diverse PS dataset, consisting of 160K images (100K for training), 274K bounding boxes, and 3K identities. It stands out from existing counterparts from two levels of diversities, \ie, scene-level and identity-level, with 92,043 scenes and significant variations in poses, clothing, scales, \etc{} for the same identity. To validate the rich context information on our dataset and make full use of it, we propose a novel global-local context network, which exploits scene and group context to boost the search performance. Extensive experiments demonstrate that MovieNet-PS is more challenging and comprehensive than existing datasets, and our approach further pushes the state of the art by a large margin (relatively 34\% in mAP) on this dataset. Our source codes, pre-trained models, and the new dataset are publicly available at: https://github.com/ZhengPeng7/GLCNet.
\end{abstract}

\section{Introduction}
\label{sec:intro}
Person search aims to match persons from natural images across different views, which simultaneously tackles pedestrian detection and person re-identification (re-ID).
In recent years, rapid progresses have been made with the renaissance of deep learning, including two-step methods~\cite{TCTS,two_stage_method1,two_stage_method2,two_stage_method3} and one-step ones~\cite{OIM,Chen2018ContextRF,Yan2021AnchorFreePS}, where the latter tends to outperform the former in terms of both accuracy and efficiency.

\begin{figure}[!t]
\begin{center}
\includegraphics[width=\linewidth]{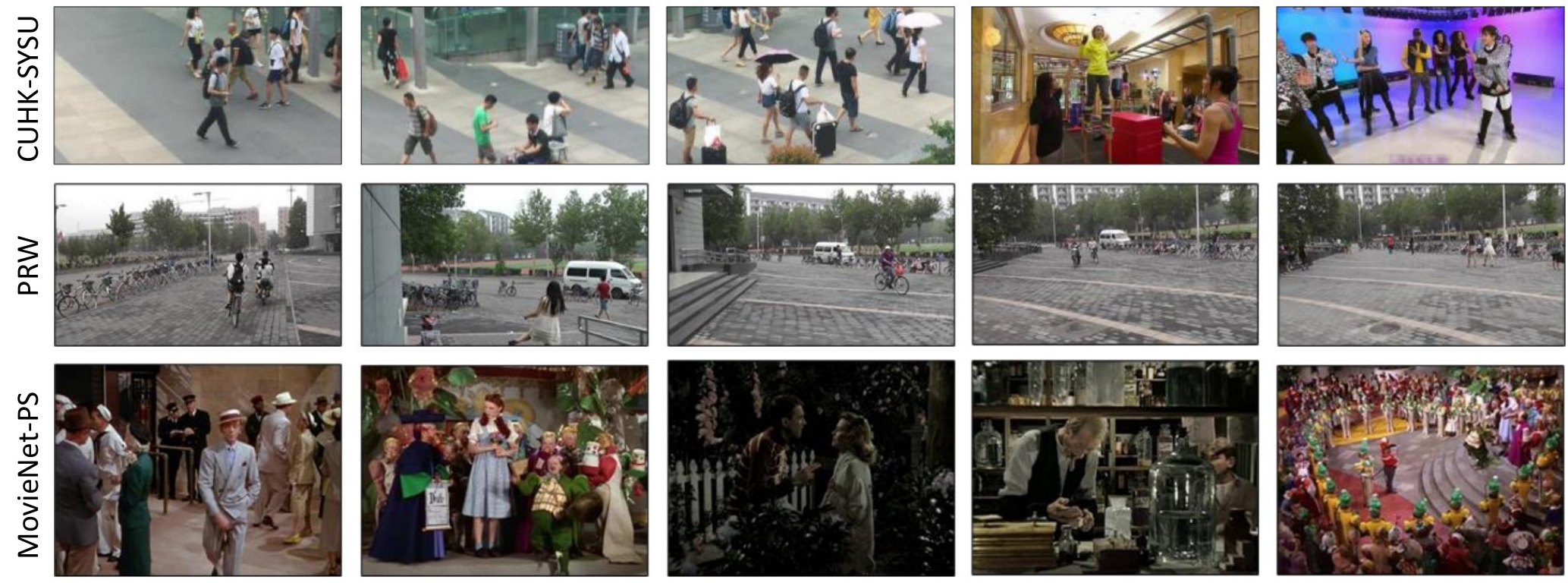}
\end{center}
\vspace{-7mm}
\caption{Comparison between different datasets. MovieNet-PS covers various scales, clothing, occlusions, and scenes, making it more challenging and closer to practical scenarios.
}
\vspace{-3mm}
\label{fig:dataset_comp}
\end{figure}

Interestingly, we notice that almost all existing methods unintentionally target at pedestrians as the query, thus restricting the application scenario within surveillance. This is probably due to that most of the current datasets (\eg, CUHK-SYSU~\cite{OIM} and PRW~\cite{PRW}) are collected with pedestrian images, aiming for intelligent visual surveillance. However, in real world, full-body images of the target person with walking/standing poses are not always available. In this case, it is still desired in some application scenarios (\eg, image search engines and social networks) to identify the target person based on diverse query images (\eg, facial images, partial body images, and portraits captured by different sensors including surveillance cameras, mobile phones, aerocameras, and \etc{}). Therefore, there is an urgent need to extend the current limited scope of person search to a more general one, \ie, searching general person images with a high degree of data diversity, mimicking the real-world scenarios.

Under this realistic setting, existing person search benchmarks have several limitations in terms of: 1) number of images (only $\sim$18K at most) ; 2) number of camera views (only 6$\sim$17) for data collection; 3) data diversity (mostly within surveillance scenarios); 4) number of instances per identity (only 5$\sim$35);
5) person actions/poses (usually involving walking/standing poses only without any event happening). The above limitations oversimplify the challenges of general person search, restricting the adaptation and generalization abilities of the models developed on these datasets.

In this paper, we fulfill the general person search task by making the following contributions. \textbf{Firstly}, we introduce a large-scale diverse person search dataset, \ie, MovieNet-PS, which is by far the largest and most diverse person search dataset. Specifically, MovieNet-PS is collected and annotated based on the large-scale MovieNet~\cite{MovieNet} dataset. We build the new dataset upon an existing one considering that collecting and annotating such a dataset is extremely expensive and privacy sensitive. In general, MovieNet-PS consists of 160K images (including a training set of 100K images), 274K bounding boxes, and 3K identities. Compared to the existing 
counterparts, MovieNet-PS exhibits its unique advantages in terms of two levels of diversities: 1) the first level of diversity lies in its 92,043 distinct scenes, covering a variety of real-world scenarios, \eg, concerts, halls, streets, and open country; 2) in terms of the same identity, the new dataset involves a unique variation in clothing, (\ie, the same person with different clothes should be identified as the same identity), and significant variations in poses, illuminations, scales, and \etc{}, as shown in \figref{fig:dataset_comp}.
All the above statistics reveal that MovieNet-PS fulfills the requirements for studying the more challenging task, \ie, general person search. \textbf{Secondly}, we propose a novel global-local context network, namely \ourmodel{}, by introducing a hybrid context learning scheme to exploit both global and local context. In \ourmodel{}, scene context and group context are facilitated for learning global discriminative features and the relationship between target persons and their neighboring co-travelers~\cite{co_traveler}, respectively. We find from the subsequent experiments that context mining is especially helpful on the new dataset, indicating the benefits of exploiting context information in realistic, highly diverse scenarios. In summary, our main contributions include:

    
    
\textbf{1)} We introduce MovieNet-PS, a new large-scale person search dataset with a high degree of data diversity, mimicking a more realistic and challenging setting for person search.

\textbf{2)} We propose \ourmodel{} by exploiting the global-local context information as an extra feature-level enhancement for the original identity features.

\textbf{3)} Extensive experimental results on MovieNet-PS show the challenging and comprehensive nature of the introduced dataset and the superiority of our context-aware method.

\begin{table*}[!t]
\centering
\resizebox{0.99\linewidth}{!}{
\begin{tabular}{l|c|c|c|c|c|c|c}
    \thickhline
    \rowcolor{mygray}  
    Datasets &   {\#Images} & {\#Boxes} & {Gallery Size} & {\#Identities}  & \makecell{\#Avg. Instances \\per Identity}  & {\#Views} & {Scenes} \\
    \hline \hline
    CUHK-SYSU~\cite{OIM}              &  18,184 & 23,430 & 50$\sim$4,000 & 8,432 & 3 & N/A & \makecell{surveillance, movies/TV} \\
    \hline
    PRW~\cite{PRW}                   &  11,816 &34,304 & 6,112 & 932 & 36 & 6 cameras & surveillance \\
    \hline
    LSPS*~\cite{LSPS}                 &  51,836 &60,433 & 33,673 & 4,067 & 15 & 17 cameras & surveillance \\
    \hline
    \textbf{MovieNet-PS}   & 160,816 & 274,274 & 2,000$\sim$10,000 & 3,087 & 10$\sim$50 & 132,115 shots & \makecell{movies} \\
    \hline
\end{tabular}
}
\vspace{-2mm}
\caption{Comparison between MovieNet-PS and other person search datasets. *LSPS is not yet publicly available.}
\label{tab:dataset_comp}
\end{table*}
\begin{table}[!t]
\centering
\resizebox{0.9\linewidth}{!}{
\begin{tabular}{c|c|c|c}
    \thickhline
    \rowcolor{mygray}
    MovieNet-PS  & {N=10} & {N=30} & {N=70} \\
    \hline\hline
    \#Identities     & 2,087  & 2,087  & 2,087 \\
    \hline
    \#Images     & 20,158  & 54,047  & 104,081 \\
    \hline
    \#Boxes     & 32,927  & 89,079  & 174,116 \\
    \hline
    \#Avg. Instances/Identity     & 10  & 26  & 50 \\
    \hline
\end{tabular}
}
\vspace{-2mm}
\caption{Different settings of the training set on MovieNet-PS. `N=10/30/70' means the maximum number of instances per identity is 10/30/70, respectively.
}
\label{tab:dataset_split}
\end{table}

\section{Related Work}
\label{sec:related}

\noindent\textbf{Person Search.}
Existing person search frameworks can be generally divided into two-step and one-step ones. In two-step frameworks, persons are detected and then fed into a re-ID model for identification, \ie, detection and re-ID models are organized in a sequential way. Inspired by object detection, one-step approaches~\cite{OIM,Yan2021AnchorFreePS,SeqNet,NAE,KD_PS,Kim_2021_CVPR} are proposed to make the joint framework more effective and efficient.
Typically, the re-ID branch in both frameworks only employs the information within the detected boxes. However, person search images naturally contain much richer information, including labeled/unlabeled persons and backgrounds, which could provide complementary information for learning more robust re-ID features, as shown in~\cite{Han_2021_ICCV, Yan2021ExploringVC}.

\noindent\textbf{Person Search Datasets.}
Xiao \etal~\cite{OIM} introduced the first person search dataset, \ie, CUHK-SYSU. It contains some images from movies/TV; however, the majority of its images come from street snaps.
PRW~\cite{PRW} is also a widely-used person search dataset collected from surveillance images.
On PRW, 11,816 surveillance frames are given with 932 identities and 6 camera views. More recently, LSPS~\cite{LSPS} was proposed as a large-scale person search dataset of surveillance frames, where many incomplete boxes were collected for improving model robustness. Although considerable progress has been achieved based on these datasets, they still have some limitations, as mentioned in Sec. \ref{sec:intro}.

\begin{figure}[!t]
  \centering
   \includegraphics[width=0.95\linewidth]{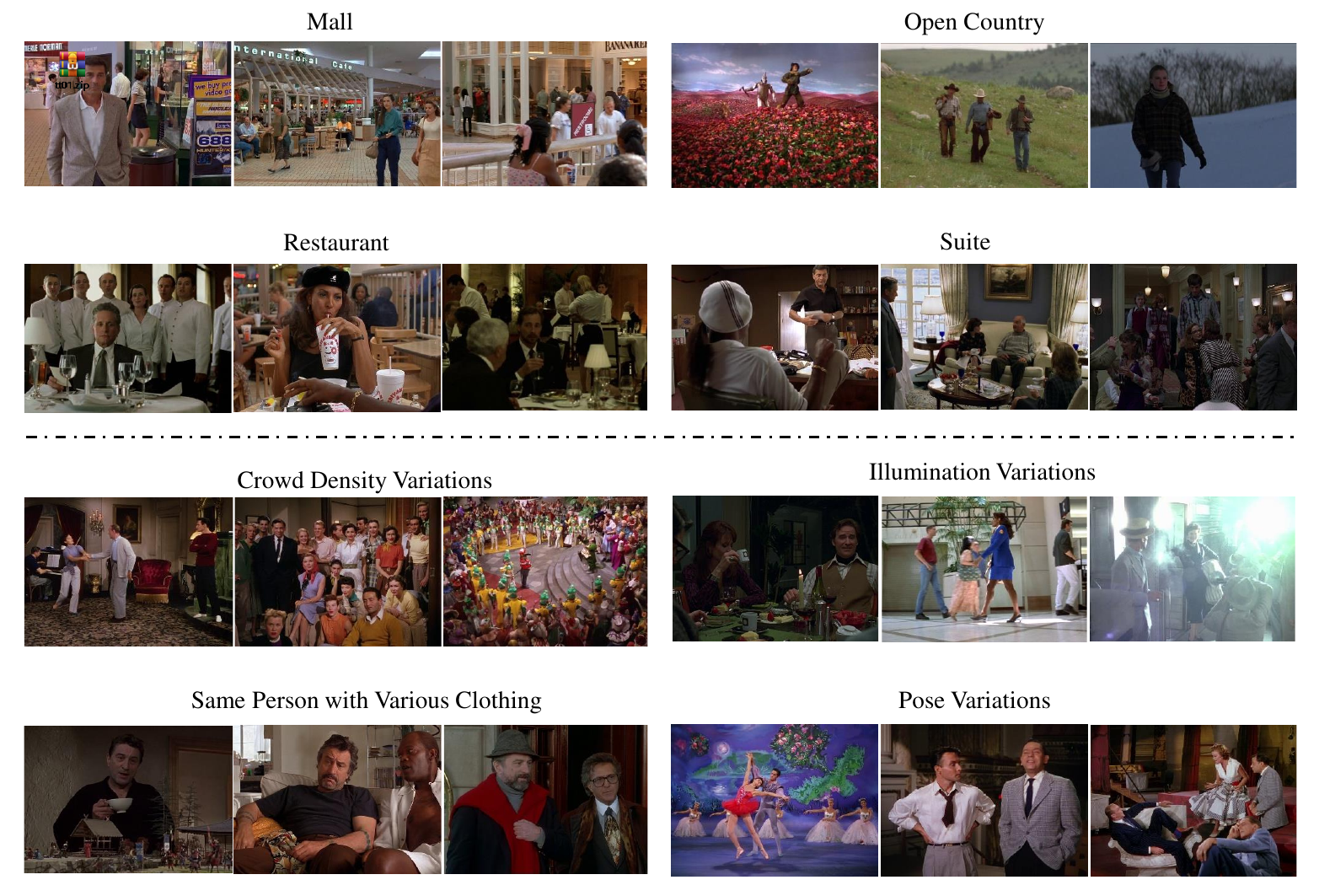}
   \vspace{-4mm}
   \caption{Typical scenarios and variations in densities, illuminations, clothing, and poses on MovieNet-PS.}
   \vspace{-4mm}
   \label{fig:dataset_diversity}
\end{figure}
\section{MovieNet-PS}
\label{sec:dataset}


We introduce a new dataset, named MovieNet-PS, containing 160K frames of 3,087 identities. Our MovieNet-PS dataset is based on the movie frames and identity labels from MovieNet~\cite{MovieNet}, which is an open-source movie dataset\footnote{http://movienet.site/}. Frames in our MovieNet-PS are collected from 385 movies of 92,043 movie shots with a large variety of backgrounds, casts and events. Since all the faces in the frames belong to famous casts who can be easily found online, there is no privacy concern. Compared to existing widely-used person search datasets, such as CUHK-SYSU~\cite{OIM} and PRW~\cite{PRW}, our introduced MovieNet-PS dataset covers a much higher degree of diversity in various ways, which makes it more challenging and closer to practical application scenarios as shown in Fig. \ref{fig:dataset_diversity} and Tab. \ref{tab:dataset_comp}. Below, we summarize the key features of our MovieNet-PS dataset.

\begin{itemize}
    \item \textbf{Extremely large data scale.} There are 160K frames on the MovieNet-PS dataset, which is 9 times (160K vs. 18K) the scale of CUHK-SYSU~\cite{OIM} and 13 times (160K vs. 12K) the scale of PRW~\cite{PRW}.

    \item \textbf{Super large number of camera views.} MovieNet-PS is collected from the movie data with 385 movies and 92,043 movie shots in total. Every single movie shot corresponds to a novel camera view.

    \item \textbf{Complicated ambient conditions.} Our movie frames are collected from movies of various genres, including actions, science fiction, suspense movies, and \etc{}~Hence, the scenarios on our dataset vary a lot.

    \item \textbf{Large variations in person clothing and poses.} MovieNet-PS dataset contains frames in countless movie plots and occasions where people may be in any state, \eg, sitting, standing, running, and dancing, where people's dresses also vary a lot. Furthermore, many casts star in different movies for different characters, so the dresses of one identity may also differ.

    \item \textbf{Abundant instances of the same identity.} Each identity on MovieNet-PS has an average number of $\sim$56 (up to 70) instances, coming from $\sim$23 different shots, which is significantly beneficial for mimicking a challenging and realistic setup for person search.
    
    \item \textbf{Multi-level data splitting.} Considering the large scale of both training and test sets, we divide the whole dataset into different subsets according to the number of training image frames and gallery size. In this way, experiments with different levels of difficulty can be conducted on MovieNet-PS.
    
    \item \textbf{Privacy insensitivity.} Since the frames are collected from movies, our dataset is not privacy concerning. Moreover, since all the facial images belong to famous casts which can be easily found on the Internet, there is no need to blur the facial regions as with the other datasets. Therefore, we expect to foster important future researches to jointly consider facial and body images for person search.
    
\end{itemize}

\begin{figure*}[t]
  \centering
   \includegraphics[width=0.95\linewidth]{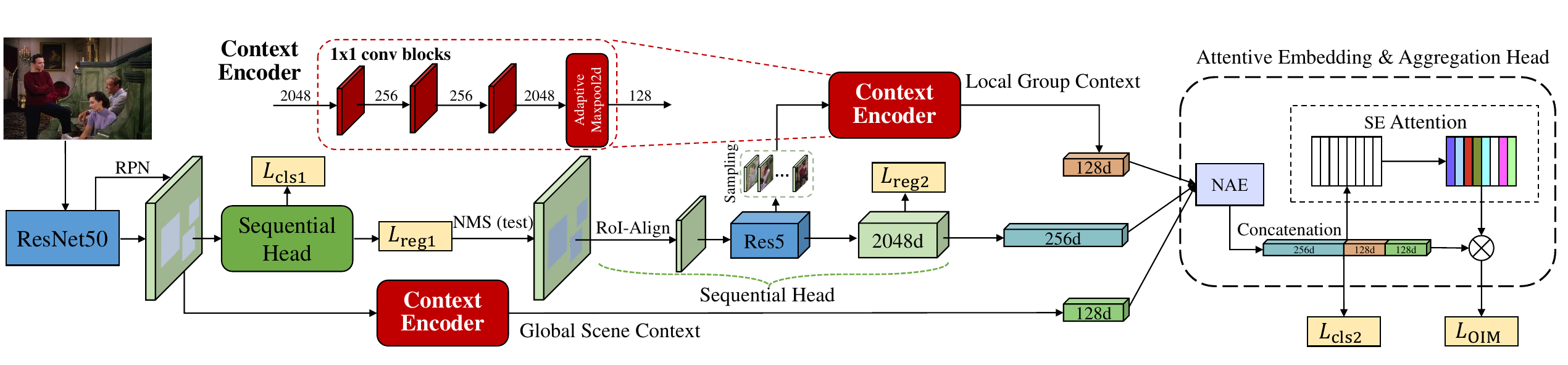}
   \vspace{-4mm}
   \caption{Overall framework of \ourmodel{}.
   We enhance the original features by simultaneously exploiting global scene and local group context. The complementary features are then fed to the attentive embedding and aggregation (AEA) head, equipped with an NAE module~\cite{NAE}, an SE Attention~\cite{SEAttention}, and an OIM loss~\cite{OIM} sequentially for learning more discriminative representations.}
   \vspace{-2mm}
   \label{fig:GLCNet}
\end{figure*}


\textbf{Evaluation Protocol.} We divide the whole dataset into training and test sets based on the 3,087 identities. 1,000 identities are included in the test set, while the rest 2,087 form the training set. However, concerning the super large scale of the whole 100K training images, we define 3 different settings for the training sets. Specifically, we adopt at most 10, 30, and 70 instances per identity, contributing to the training sets of 20K, 54K, and 100K training images, respectively. Following CUHK-SYSU~\cite{OIM}, we introduce different gallery sizes to bring various levels of difficulties for comprehensive evaluation. For more details about each subset of the training set, please refer to Tab. \ref{tab:dataset_split}.
Under such divisions of training and test sets, we can train and evaluate the model or learning strategy at different levels. As for the evaluation metrics, we adopt mean average precision (mAP) and top-1 accuracy for person search, and recall and average precision (AP) for detection.

\section{Global-Local Context Network}
\label{sec:method}

Fig. \ref{fig:GLCNet} illustrates the overall framework of \ourmodel{}. We adopt the recently proposed SeqNet~\cite{SeqNet} as our base network due to its concise and effective architecture, \ie, the sequential pipeline in the middle of Fig. \ref{fig:GLCNet}. The base network combines the sub-networks of detection and re-ID in a sequential and end-to-end manner. Built upon this base network, our goal is to fully exploit the rich context existing in the input image. Unlike previous works~\cite{Yan2019LearningCG,group_reID_cxt1} that develop sophisticated schemes to capture the context information, we simply combine features of both target and context as the new re-ID features. Specifically, global scene context (GSC) features are encoded with a context encoder, based on the feature map of the entire input image. Local group context (LGC) features are obtained from the RoI-Aligned person-level features through a context encoder of the same architecture as above. Finally, an attentive embedding and aggregation (AEA) head is developed to further augment the above different levels of features to output the final discriminative features.

As shown in Fig. \ref{fig:GLCNet}, AEA receives the global scene and local group contextual information and the typically used target person features. These features of different levels are first fed to an NAE layer~\cite{NAE} for feature calibration, and then aggregated to form a unified feature. As a result, the identity features of a certain person consist of three parts: its original re-ID features, global context features that perceive the scene, and local context features representing its surrounding persons. Subsequently, the attention mechanism~\cite{SEAttention} is applied to the unified feature to focus on the most informative channels. Finally, we learn the final representation of the target person through optimizing the OIM loss same as~\cite{OIM,NAE,SeqNet}.

\begin{table}[!t]
\centering
\resizebox{0.85\linewidth}{!}{
\begin{tabular}{c|c|c|c|c|c}
    \thickhline
    \rowcolor{mygray}
    Method  & {OIM} & {NAE} & {NAE+}    & {SeqNet}  & {\textbf{\ourmodel{}}}    \\
    \hline\hline
    mAP     & 27.5  & 30.5  & 31.2      & 34.2      & \textbf{45.9}      \\
    \hline
    top-1   & 70.1  & 73.8  & 75.4      & 80.9      & \textbf{85.6}      \\
    \hline
\end{tabular}
}
\vspace{-1.5mm}
\caption{Performance of different methods under the setting of `Training Set:N=10'-`Gallery Size:2K'.}
\label{tab:ps100k_n10gs2k}
\end{table}
\begin{table}[!t]
\centering
\resizebox{0.99\linewidth}{!}{
\begin{tabular}{c|c|cc|cc|cc}
    \thickhline
    \rowcolor{mygray}
    & &\multicolumn{2}{c|}{N=10}  & \multicolumn{2}{c|}{N=30} & \multicolumn{2}{c}{N=70}\\
    \cline{3-8}
   \rowcolor{mygray}
   \multirow{-2}{*}{Gallery} & \multirow{-2}{*}{Method} & \multicolumn{1}{c}{mAP} & \multicolumn{1}{c|}{top-1} & \multicolumn{1}{c}{mAP} & \multicolumn{1}{c|}{top-1} & \multicolumn{1}{c}{mAP} & \multicolumn{1}{c}{top-1}\\
    \hline\hline
    & SeqNet & 34.2 & 80.9 & 41.6 & 85.5 & 44.0 & 85.6 \\
    \multirow{-2}{*}{2K}& \textbf{\ourmodel{}} & \textbf{45.9} & \textbf{85.6} & \textbf{53.6} & \textbf{89.4} & \textbf{55.5} & \textbf{88.6} \\
    \hline
    & SeqNet & 30.0 & 79.9 & 37.6 & 84.4 & 39.0 & 85.7 \\
    \multirow{-2}{*}{4K}&\textbf{\ourmodel{}} & \textbf{41.5} & \textbf{82.4} & \textbf{50.0} & \textbf{87.6} & \textbf{51.1} & \textbf{87.8} \\
    \hline
     & SeqNet    & 24.1 & 77.3 & 30.9 & 81.4 & 32.6 & 82.4 \\
    \multirow{-2}{*}{10K}&\textbf{\ourmodel{}}& \textbf{34.4} & \textbf{78.5} & \textbf{42.7} & \textbf{84.7} & \textbf{44.4} & \textbf{85.7} \\
    \hline
\end{tabular}
}
\vspace{-1.5mm}
\caption{Performance comparison in terms of different splits.
}
\label{tab:ps100k_bsl}
\vspace{-2mm}
\end{table}

\textbf{Implementation Details.} We implement our model using PyTorch. All experiments are conducted on a single NVIDIA Tesla V100 GPU. During training, the batch size is set to 15 and each image is resized to 720$\times$240. We set the circular queue size of OIM~\cite{OIM} to 3,000. The only data augmentation used is random horizontal flip. For simplicity, the losses used in our framework are identical to those in previous works~\cite{OIM,SeqNet,NAE}, including OIM~\cite{OIM} ($L_{\mathrm{OIM}}$), regression ($L_{\mathrm{reg1}}$ and $L_{\mathrm{reg2}}$), and classification ($L_{\mathrm{cls1}}$ and $L_{\mathrm{cls2}}$) losses for re-id and detection (as shown in \figref{fig:GLCNet}). We follow SeqNet~\cite{SeqNet} to set all the hyper-parameters during training/inference, including learning rates, IoU thresholds, CBGM~\cite{SeqNet}, and \etc{}

\section{Experiments}
\label{sec:exp}
We report the performance of different methods under a typical setting in Tab. \ref{tab:ps100k_n10gs2k}, where we can observe the significant improvement of \ourmodel{} over other competitors, especially in mAP (\eg, 34\% relatively higher than the second-best SeqNet). Besides, since we have 3 levels of training sets and 3 gallery sets, we conduct comparisons between SeqNet and our \ourmodel{} under all 9 settings. As shown in Tab. \ref{tab:ps100k_bsl}, \ourmodel{} consistently outperforms SeqNet in terms of all settings. The effect of each component, \ie{}, base, scene context, and group context is provided in \tabref{tab:ablation}.
The above results on such a highly diverse dataset verifies the effectiveness of exploiting both global and local context, while other person search approaches cannot perform well in the challenging setting.


\begin{table}[!t]
\centering
\resizebox{0.7\linewidth}{!}{
\begin{tabular}{ccc|cc}
\thickhline
\rowcolor{mygray}  
 {Baseline}                  &  {Scene}                  &  {Group}                & mAP                  & \multicolumn{1}{c}{top-1}                  \\ 
\hline \hline  
$\checkmark$ &   &  & 34.6 & 80.8                \\
$\checkmark$ & $\checkmark$  &  & 42.5 & 83.4              \\
 $\checkmark$ &  & $\checkmark$& 38.6 & 82.2                        \\
$\checkmark$ & $\checkmark$  & $\checkmark$ & \textbf{45.9} & \textbf{85.6}          \\\hline
\end{tabular}
}
\vspace{-1.5mm}
\caption{Comparative results with different context features under the setting of `Training Set:N=10'-`Gallery Size:2K'.}
\label{tab:ablation}
\end{table}
\begin{table}[!t]
\centering
\resizebox{0.99\linewidth}{!}{
\begin{tabular}{c|cc|cc|cc}
\hline\thickhline
\rowcolor{mygray}  
 & \multicolumn{2}{c|}{MovieNet-PS}    & \multicolumn{2}{c|}{CUHK-SYSU}    & \multicolumn{2}{c}{PRW}                        \\ \cline{2-7} 
\rowcolor{mygray}  
\multirow{-2}{*}{Test \textbackslash Train} & mAP                  & \multicolumn{1}{c|}{top-1}  & mAP                  & \multicolumn{1}{c|}{top-1}  & mAP                  & \multicolumn{1}{c}{top-1} \\ 
    \hline\hline
    MovieNet-PS     & - & - & 0.7 & 5.6  & 4.6 & 28.5 \\
    \hline
    CUHK-SYSU     & 41.5 & 42.1 & - & - & 54.3 & 56.4 \\
    \hline
    PRW     & 4.5 & 31.0  & 25.3 & 77.0  & - & - \\
    \hline
\end{tabular}
}
\vspace{-1.5mm}
\caption{Cross-dataset experimental results.}
\label{tab:generalization}
\vspace{-2mm}
\end{table}

\textbf{Cross-Dataset Evaluation.}
\label{sec:gen_abi}
To quantitatively demonstrate the gap between MovieNet-PS and other person search datasets, we conduct cross-dataset experiments based on a direct transfer scheme, \ie, training our \ourmodel{} on one dataset and then directly testing it on other datasets.
From \tabref{tab:generalization}, we can see the model trained on MovieNet-PS can achieve 41.5\% and 4.5\% in mAP on CUHK-SYSU and PRW, respectively, while the models trained on CUHK-SYSU and PRW can only achieve 0.7\% and 4.6\% in mAP on MovieNet-PS, respectively. This again verifies that MovieNet-PS is much more challenging than the existing counterparts.

\section{Conclusion}
In this paper, we propose the large-scale person search dataset and the context-aware person search framework that exploits both global scene context and local group context in a unified manner. Qualitative analysis and quantitative results demonstrate that the new dataset brings much more challenges to the community and our context-aware method has a strong ability to tackle the person search task under this realistic setting.

\bibliographystyle{IEEEbib}
\bibliography{refs}

\end{document}